\documentclass[10pt,twocolumn,letterpaper]{article}

\usepackage{cvpr}              % To produce the CAMERA-READY version
\usepackage{makecell}
\usepackage{arydshln}
\usepackage{cellspace}
\usepackage{subcaption}

\usepackage{times}
\usepackage{latexsym}
\usepackage{pifont}
\usepackage{amsmath}
\usepackage{multirow}
\usepackage{booktabs} 
\usepackage{amssymb}
\usepackage{tablefootnote}
\usepackage{colortbl} %\cellcolor
\usepackage{paralist} %\begin{compactitem}
\usepackage{soul} %\sethlcolor
\usepackage{xspace}
\usepackage{tabularx}
\usepackage{graphicx}
\newcommand{\ourmodel}{PerceptionGPT\xspace}
\newcommand{\loctoken}{\textit{$<$vis$>$\xspace} }
\newcommand{\cmark}{\scalebox{0.8}{\ding{52}}}
\newcommand{\xmark}{\ding{55}}  % cross mark
\definecolor{cvprblue}{rgb}{0.21,0.49,0.74}
\usepackage[pagebackref,breaklinks,colorlinks,citecolor=cvprblue]{hyperref}

\def\paperID{*****} % *** Enter the Paper ID here
\def\confName{CVPR}
\def\confYear{2024}

\title{PerceptionGPT: Effectively Fusing  Visual Perception into LLM}

\author{
Renjie Pi$^1$
\quad Lewei Yao$^1$%\footnotemark[2]
\quad Jiahui Gao$^2$
\quad Jipeng Zhang$^1$
\quad Tong Zhang$^1$ \\
$^1$The Hong Kong University of Science and Technology \\
$^2$The University of Hong Kong  \\
}

\begin{document}
\maketitle
\begin{abstract}
The integration of visual inputs with large language models (LLMs) has led to remarkable advancements in multi-modal capabilities, giving rise to visual large language models (VLLMs). However, effectively harnessing VLLMs for intricate visual perception tasks remains a challenge. 
% Existing approaches can be categorized into two groups: (1) two-stage-based approaches that borrow the power from external vision experts, which excel at visual tasks but unable to make LLM interpret spatial information, and (2) end-to-end approaches that enables the model to perform fine-grained visual perception tasks on its own, but encounters difficulties during training, suffer from quantization errors and struggle to handle a large number of bounding boxes or segmentation masks due to the lengthy sequence involved. 
In this paper, we present a novel end-to-end framework named \textbf{\ourmodel}, which efficiently and effectively equips the model with visual perception abilities by leveraging the representation power of LLMs' token embedding. Our proposed method treats the token embedding of the LLM as the carrier of visual information, then leverage lightweight visual task encoders and decoders to handle the visual perception signals (e.g., bounding boxes, segmentation masks). Our approach significantly alleviates the training difficulty suffered by previous approaches that formulate the perception signals as discrete tokens, and enables achieving superior performance with fewer trainable parameters, less training data and shorted training time. Moreover, as only one token embedding is required to represent the perception signal, the resulting sequence length during inference is significantly reduced. Consequently, our approach enables accurate and flexible visual representations, and efficient handling of a complex perception signals. We validate the effectiveness and efficiency of our approach through extensive experiments. The results demonstrate significant improvements over previous methods with much fewer trainable parameters and GPU hours.
% , which facilitates future research in enabling LLMs with visual perception abilities.
\end{abstract}    
\section{Introduction}
\label{sec:intro}
The rapid advancements in deep learning and natural language processing have given rise to large language models (LLMs) capable of comprehending and generating human-like text~\cite{brown2020language, scao2022bloom, chowdhery2022palm, smith2022using, ouyang2022training, bai2022training, touvron2023llama, vicuna2023}. Recently, the development of visual large language models (VLLMs), which combine visual inputs with LLMs, has demonstrated impressive multi-modal capabilities and opened up new possibilities beyond text-based tasks~\cite{cho2021unifying, openai2023gpt4, liu2023llava, zhu2023minigpt4, su2023pandagpt, bai2023qwenvl}.
\begin{figure}[t]
        \centering
        \includegraphics[width=1.0\linewidth]{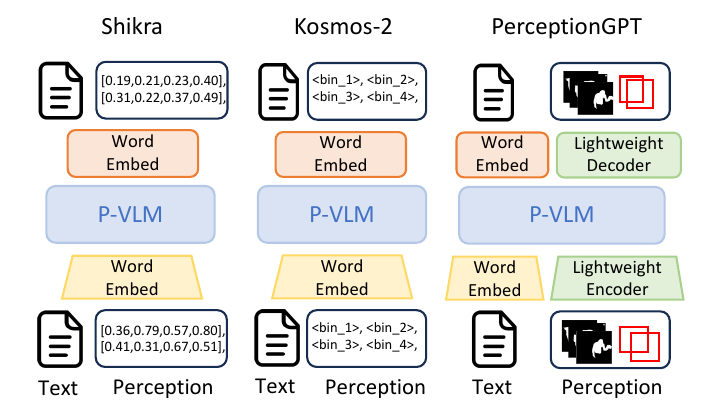}
    %  \vspace{-0.5cm}
    \caption{Illustration of different strategies to encode and decode visual perception information. Previous approaches formulate the visual information into discrete tokens in the same way as text. On the other hand, our \ourmodel leverages lightweight visual encoder and decoders to fuse such information into the embedding space of LLM.}
     \vspace{-0.33cm}
    \label{fig:framework}
\end{figure}

However, enabling VLLMs to perform complex visual perception tasks, such as object detection and segmentation, remains a significant challenge. Current state-of-the-art approaches can be divided into two categories: 1) two-stage-based approaches that leverage a vision expert alongside the reasoning ability of the LLM to handle visual perception tasks (e.g., ViperGPT~\cite{surís2023vipergpt}, DetGPT~\cite{pi2023detgpt}, MMReact~\cite{yang2023mmreact}, and LISA~\cite{lai2023lisa}). While these approaches can utilize vision experts to excel at visual tasks, the LLM itself lacks the ability to interpret visual perception inputs, thereby limiting their applicability and flexibility. 2) End-to-end approaches that integrate visual perception capabilities into the LLM (e.g., Kosmos-2~\cite{peng2023kosmos2}, Shikra~\cite{chen2023shikra}, VisionLLM~\cite{wang2023visionllm}, and Qwen-VL~\cite{bai2023qwenvl}). We refer to these as perception-enhanced vision-language models (P-VLMs). These approaches enable the input and output of object boxes or segmentation masks, allowing the LLM to interpret visual inputs. Such capabilities have the potential to empower various applications, including intelligent robotics~\cite{ahn2022i, brohan2023rt1}, autonomous driving~\cite{Yurtsever_2020, ding2023hilm}, and virtual reality~\cite{HUYNHTHE2023105581}.

% since the visual perception information do not naturally possess the causal structure as in language, decoding them in a causal manner poses more challenges during training. 
Despite the promising abilities of P-VLMs, they predominantly represent visual perception signals (e.g., bounding boxes, segmentation masks) as sequences of discrete tokens, which introduces several weaknesses: 1) Since the perception signals are always in continuous forms, it is not natural to model them as sequences  of discrete tokens. As a result, it becomes difficult to train, which requires more data, more trainable parameters and training steps to enable good performance. For example, Shikra~\cite{chen2023shikra} requires 120 hours to train on 8 A100 GPUs with 80G memory, while Kosmos-2\cite{peng2023kosmos2} takes 1 day to train on 256 V100 GPUs. Such training cost is prohibitive for most researchers; 2) the discretization of perception signals inevitably introduces quantization error, potentially causing accuracy drop; 3) representing visual perception as sequences of discrete tokens introduces redundant tokens and require a long context window, which becomes difficult to handle when the signal is complex ( e.g., segmentation masks).

In this paper, we propose \textbf{\ourmodel}, a novel framework that effectively integrates visual perception capabilities into P-VLMs. Our core intuition is that the high-dimensional token embedding contains rich information, which should be able to carry the essential information to represent a perception signal. However, prior methods, which relied on representing these signals through discrete tokens, fail to harness the strong representational strength of these embeddings. To address this, we propose to leverage the LLM's token embedding as the carrier of information to represent the perception signal. Specifically, we introduce a special token \loctoken, this token serves as a marker for the presence of a perception signal within the given context, distinct from standard word tokens. Unlike ordinary tokens that represent single words, the \loctoken's embedding can encompass a diverse range of semantics and information.  For instance, segmentation masks of various shapes can be encoded into the embedding through a lightweight encoder. Correspondingly, the embedding of \loctoken can be decoded back into these masks, whose shapes are dependent on the preceding context, via a lightweight decoder.
% from its token embedding, the task-specific visual outputs can be then be decoded via lightweight visual perception decoders.

% During training, a combination of auto-regressive language modeling loss and task-specific objectives specifically designed for vision tasks (e.g., GIoU loss for bounding box, DICE loss for segmentation masks) can be employed in conjunction. The language modeling loss enables the LLM to generate responses to user-input, while determine when to produce visual signals by generating \loctoken tokens, while the task-specific losses equip the model with visual perception abilities.

During training, a combination of auto-regressive language modeling loss and task-specific objectives specifically designed for vision tasks (e.g., GIoU loss for bounding box, DICE loss for segmentation masks) can be employed in conjunction. The language modeling loss enables the Large Language Model (LLM) to generate responses to user inputs and to decide when to produce visual outputs through the generation of \loctoken tokens. Concurrently, the task-specific losses empower the model with enhanced visual perception capabilities. These losses ensure that the model is equipped with both linguistic fluency and visual perception acuity.
% Instead of representing visual perception as discrete tokens as in conventional approaches, we utilize the LLM's token embedding to encode such information. 

Our method offers several advantages: (1) leveraging the token embeddings to represent perception signals greatly alleviates the training difficulty. As a result, \ourmodel obtains promising results by tuning only a small fraction of parameters (e.g., LoRA), where previous approaches completely fail or suffer dramatic performance degradation (Table~\ref{tab:tokenization}); (2) our approach allows for more accurate representations of visual perception by predicting the exact values, effectively addressing the quantization errors inherent in discrete token formulations; (3) compared with previous approaches, we only require a single special token \loctoken to represent the perception signal (Table~\ref{tab:token_number}), resulting in reduced context length which significantly accelerate the decoding process. Specifically, we choose two of the most representative visual perception tasks, i.e., detection and segmentation, to showcase the effectiveness of our proposed method, while other perception tasks such as depth or pose estimation can be easily integrated.

% This paper elaborates the details of our proposed \ourmodel framework, including the architectural design and training procedures necessary to incorporate language modeling and task-specific objectives. 
We summarize the contributions of our paper as follows:
% \begin{itemize}
% \setlength{\itemsep}{0pt}
% \setlength{\parsep}{0pt}
% \setlength{\parskip}{0pt}
%     \item We propose a practical approach named \ourmodel for tackling the heterogeneity problem, which extracts and exploits the hidden information from the global model's trajectory. In this way, the server can access the essential knowledge of the global data distribution to reinforce aggregation;
%     \item We experimentally show the synthesized dataset helps stabilize training, boost convergence and achieve 
%     % better performance
%     significant performance improvement
%     under heterogeneity;
%     \item We provide insights and detailed analysis into the working mechanisms of the proposed \ourmodel both experimentally and theoretically.
%     % \prj{\cite{arpit2017closer}, early rounds, common patterns, efficient}.
% \end{itemize}
\begin{itemize}
\setlength{\itemsep}{0pt}
\setlength{\parsep}{0pt}
\setlength{\parskip}{0pt}
    \item We propose a general framework for training perception-enhanced vision language model (P-VLM) by taking advantage of the representation power of LLM's token embedding, which enables achieving promising performances in parameter-effcient and data-efficient manner.
    \item Our approach eliminates the error introduced by discretization, which further boosts the performance. In addition, we require only one token to encode the visual perception information, which greatly reduces the sequence length and improves inference efficiency.
    \item We conduct extensive experiments on various benchmarks, including referring expression comprehension (REC), referring expression segmentation (RES), image captioning (IC) and visual question answering (VQA) to validate the effectiveness and efficiency of \ourmodel. Notably, we achieve SOTA performances on REC and RES tasks with only 300M tunable parameters, which is around 4\% of previous approach for P-VLM.
\end{itemize}

% \item 

\begin{figure*}
\includegraphics[width=1.0\textwidth]{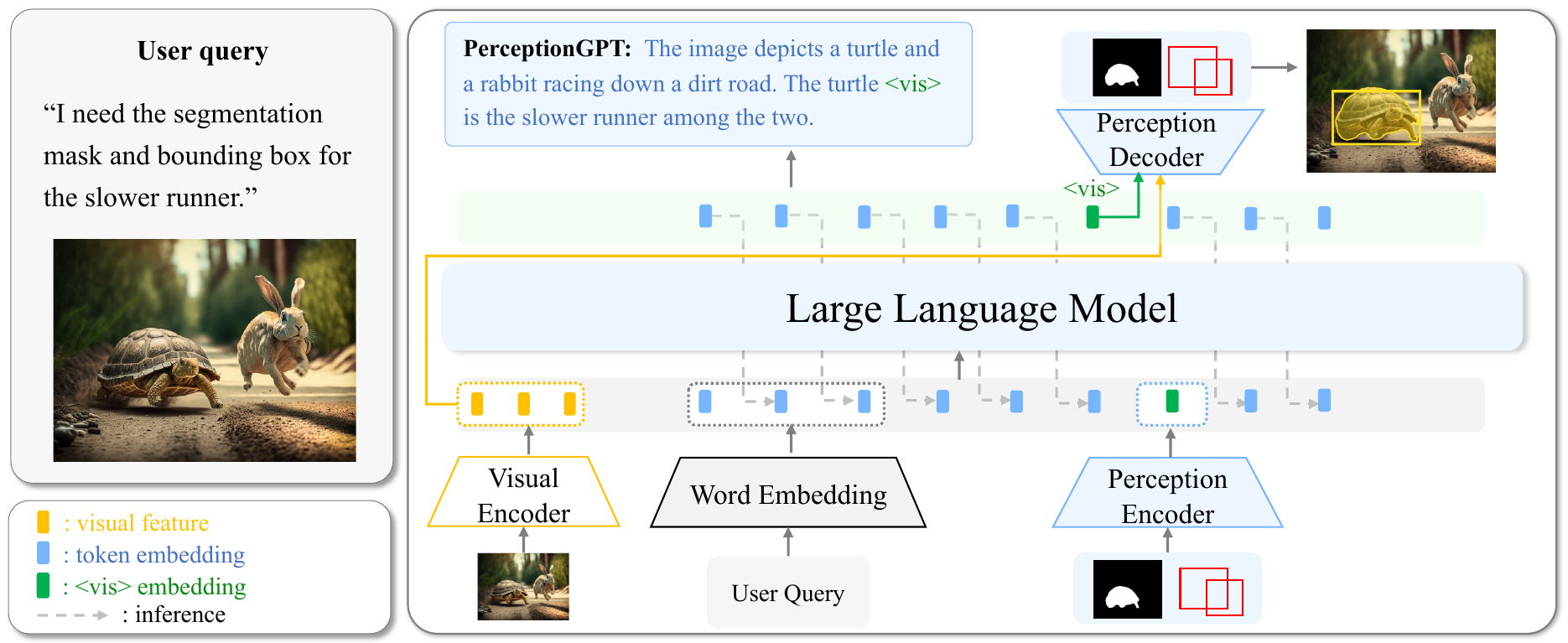} 
\caption{ The illustration of framework of PerceptionGPT. The model is trained in an end-to-end manner, rather than outputting the location and coordinates in the form of discrete tokens, each box and mask can be represented by one single continuous embedding from the LLM's output. Both the perception encoder and decoder are lightweight architectures (e.g., MLP, ResNet) trained from scratch, without relying on any pretrained visual experts.
% During the process, the LLM acts as the brain, while the detector empowers the system with the ability to "see". 
}\label{fig:framework}
\end{figure*}
\begin{table*}
\centering
\begin{tabular}{|l|l|c|c|c|c|c|c}
\toprule
% \multirow{2}{*}{Model} & \multicolumn{2}{c|}{Image Cap.} & \multicolumn{2}{c|}{Image Cap.} & End-to-End \\
% & \& Rea. & Multi-Region & \& Rea. & Multi-Round & Model \\
\multicolumn{2}{c|}{\multirow{2}{*}{Model}} & Image & Region & Bounding
& Segmentation & Multi-Instance & End-to-End \\
&& Caption & Caption & Box & & Segmentation & \\
\hline
\multirow{3}{*}{Two Stage}&Visual ChatGPT~\citep{wu2023visual} & \cmark & \xmark & \cmark & \cmark & \cmark & \xmark \\
\cline{2-8}
&DetGPT~\citep{pi2023detgpt} & \cmark & \xmark & \cmark & \xmark & \xmark & \xmark \\
\cline{2-8}
& LISA~\cite{lai2023lisa} & \cmark & \xmark & \xmark & \cmark & \xmark & \xmark \\
\cline{2-8}
\hline
\multirow{8}{*}{End-to-End} &MiniGPT-4~\cite{zhu2023minigpt4} & \cmark & \xmark & \xmark & \xmark & \xmark & \cmark \\
\cline{2-8}
&LLaVA~\citep{liu2023llava} & \cmark & \xmark & \xmark & \xmark & \xmark & \cmark\\
\cline{2-8}
&InstructBLIP~\citep{dai2023instructblip} & \cmark & \xmark & \xmark & \xmark & \xmark & \cmark\\
% \hline
% MM-REACT & \cmark & \cmark & \cmark & \cmark & \xmark \\
% \hline
% InternGPT & \cmark & \xmark & \xmark & \xmark & \xmark \\
\cline{2-8}
&VisionLLM~\citep{wang2023visionllm} & \cmark & \xmark & \cmark & \cmark & \cmark & \cmark\\
\cline{2-8}
% CaptionAnything & \cmark & \xmark & \cmark & \xmark & \xmark \\
% \hline
&GPT4RoI~\citep{zhang2023gpt4roi} & \cmark & \cmark & \xmark &  \xmark & \xmark & \cmark\\
\cline{2-8}
&Shikra~\cite{chen2023shikra} & \cmark & \cmark & \cmark & \xmark & \xmark & \cmark\\
\cline{2-8}
&Kosmos-2~\cite{peng2023kosmos2} & \cmark & \cmark & \cmark & \xmark & \xmark & \cmark\\
\cline{2-8}
&\textbf{\ourmodel} & \cmark &  \cmark & \cmark & \cmark & \cmark & \cmark\\
\bottomrule
\end{tabular}
\caption{Comparisons of functionalities supported by different vision-language models. Our \ourmodel is an end-to-end model that supports image-level and region-level understanding, object localization and segmentation tasks.}
\end{table*}

\section{Related works}
\subsection{Vision Large Language Models}
In recent years, significant progress has been made in large language models like GPT-3~\citep{brown2020language}, Bloom~\citep{scao2022bloom}, PaLM~\citep{chowdhery2022palm}, megatron-turing-530b~\citep{smith2022using}, Chinchilla~\citep{hoffmann2022training}, InstructGPT~\citep{ouyang2022training}, LLaMA~\citep{touvron2023llama} and Anthropic-LM~\citep{bai2022training}, pushing the boundaries of language understanding and generation. These models have demonstrated human-level abilities in various tasks. The success of language models has also driven research on vision-language interaction, resulting in the development of various multi-modal models. Recent works, such as BLIP2~\cite{li2023blip2}, LLaVA~\citep{liu2023llava}, MiniGPT4~\citep{zhu2023minigpt4}, InstructBlip~\cite{dai2023instructblip} and GPT-4~\citep{openai2023gpt4}, have demonstrated the potential of multimodal interaction, leveraging the powerful capability of LLMs to build VLLMs.

\begin{figure*}
\includegraphics[width=1.0\textwidth]{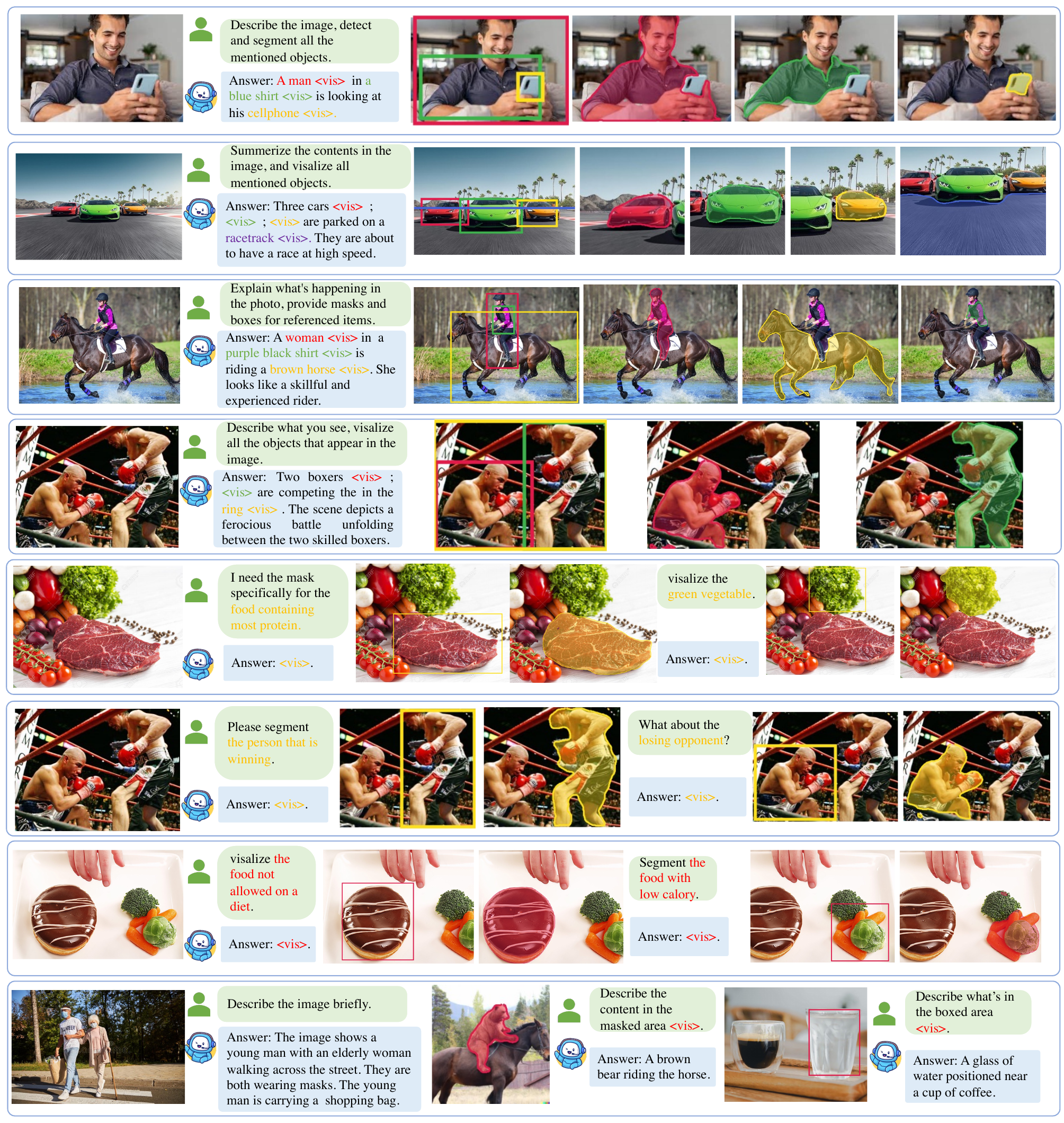} 
\caption{ Visualization of results from \ourmodel. Our proposed framework enables effectively fusing visual perception capability into P-VLM while maintaining its generation and reasoning ability. Row [1-4], row [5-8] demonstrate spot captioning and reasoning segmentation/detection, respectively. row [10] demonstrates image-level captioning and region-level captioning.
% During the process, the LLM acts as the brain, while the detector empowers the system with the ability to "see". 
}\label{fig:perceptiongpt_demo}
\end{figure*}
\subsection{Two-stage Vision Language Assistant} Recent research trends merge LLMs with vision expert models for tasks needing reasoning. Visual ChatGPT~\citep{wu2023visual, yang2023mmreact} uses LLMs as planners for visual expert APIs. ViperGPT~\citep{surís2023vipergpt} generates code with LLMs to call visual expert functions. DetGPT~\citep{pi2023detgpt} applies VLLMs for instruction interpretation and object identification using external open-vocabulary detectors~\cite{yao2022detclip, li2022grounded,liu2023grounding, yao2021joint, yao2021gdetkd}. LISA~\citep{lai2023lisa} combines VLLMs with the Segmentation Anything Model (SAM)~\citep{kirillov2023segment} for segmentation mask creation. Although those methods perform well on visual perception tasks, they require an external vision expert,  which limits their applicability for tasks that lack such expert models. In addition, the VLLM still has no capability of understanding the perception signal. 

\subsection{Perception-enhanced Vision Language Model}  More recently, works such as VisionLLM~\citep{wang2023visionllm}, Kosmos-2\citep{peng2023kosmos2} and Shikra~\citep{chen2023shikra} have made the initial attempt to integrate localization capability into VLLMs, which have demonstrated great performance and open up a series of new possibilities. These model represent the perception information as a series of discrete tokens, where each token stands for a 2-D coordinate.  Specifically, ~\citep{wang2023visionllm, peng2023kosmos2} introduce new tokens into the LLM to represent the 2D coordinates, while ~\citep{chen2023shikra} directly use numbers to represent the bounding boxes,  which improves accuracy at the cost of longer sequence lengths and slower inference. Despite thee promising results, such discrete formulation faces disadvantages such as difficulty in training, quantization error due to the discrete representation, and redundant tokens that makes the context length prohibitive if the number of masks/boxes are large. On the other hand, we propose a method to embed the perception information into a single token embedding, which addresses the above issues and provides a solution for training a strong P-VLM efficiently.

\begin{table*}[htp!]
% \small
    \centering
 % \vspace{-0.25cm}
    % \resizebox{\textwidth}{!}{%
% \begin{small}
% \scalebox{0.9}
{
\begin{tabular}{c!{\vrule width 0.5pt}c!{\vrule width 0.5pt}ccc!{\vrule width 0.5pt}ccc!{\vrule width 0.5pt}cc}
% \begin{tabular}{p{1.3cm}!{\vrule width 0.5pt}p{1.3cm}p{1.3cm}p{1.3cm}!{\vrule width 0.5pt}p{1.3cm}p{1.3cm}p{1.3cm}!{\vrule width 0.5pt}p{1.3cm}p{1.3cm}p{1.3cm}}

\toprule
 & &        \multicolumn{3}{c}{RefCOCO} & \multicolumn{3}{c}{RefCOCO+} & \multicolumn{2}{c}{RefCOCOg}\\
 % \cline{2-3} \cline{4-6} \cline{7-9}
Model Type &Method & val &  testA & testB &  val &  testA & testB &  val & test \\
\midrule
\multirow{9}{*}{Generalist VL SOTAs} &GPV-2~\citep{kamath2022webly}     & 51.59  & - & - & - & -  & - & - & - \\
&OFA-L~\citep{wang2022ofa}    &   79.96 &  83.67 &  76.39 & 68.29 & 76.00 & 61.75 & 67.57  & 67.58 \\
&Unified-IO~\citep{lu2022unifiedio}    &  78.60  & - & - & - & -  & - & - & - \\
% \midrule
&OFASys~\citep{bai2022ofasys}     &  - & 80.10 & - & - & - & - & - & - \\
&VisionLLM-H~\citep{wang2023visionllm}    &  - & 86.70 & - & - & - & - & - & - \\
&Shikra-7B~\citep{chen2023shikra}    &  87.01 & 90.61 & 80.24 & 81.60 & 87.25 & 73.20 & 82.27 & 82.19\\
&Shikra-13B~\citep{chen2023shikra}    &  87.83 & 91.11 & 81.81 & 82.89 & 87.79 & 74.41 & 82.64 & 83.16\\
&\ourmodel-7B    &  \textbf{88.59} & \textbf{92.51} & \textbf{84.60} & \textbf{82.05} & \textbf{88.60} & \textbf{74.21} & \textbf{84.13} & \textbf{85.20}\\
&\ourmodel-13B    &  \textbf{89.17} & \textbf{93.20} & \textbf{85.96} & \textbf{83.72} & \textbf{89.19} & \textbf{75.31} & \textbf{83.75} & \textbf{84.69}\\
\midrule
\multirow{2}{*}{Specialist SOTAs}&TransVG~\citep{deng2022transvg} & 80.83 & 83.38 & 76.94 & 68.00 & 72.46 & 59.24 & 68.03 & 68.71\\
&SeqTR~\citep{zhu2022seqtr} & 83.72 & 86.51 & 81.24 & 71.45 & 76.26 & 64.88 & 74.86 & 74.21\\
% &G-DINO-L~\citep{liu2023grounding} & 90.56 & 93.19 & 88.24 & 82.75 & 88.95 & 75.92 & 86.13 & 87.02\\
&UNINEXT-H~\citep{yan2023universal} & 92.64 & 94.33 & 91.46 & 85.24 & 89.63 & 79.79 & 88.73 & 89.37\\
\bottomrule
\end{tabular}
}
\caption{Results on referring expression comprehension (REC). We compare our  \ourmodel with both the generalist models and specialist models. Our method achieves outstanding performance on REC tasks with only parameter-efficient (LoRA) training. }
\label{tab:ref_det}
% \end{small}
% }
% \vspace{-0.35cm}
\end{table*}
\section{Method}
\label{sec:method}
In this section, we introduce our proposed \ourmodel  framework, which presents an efficient and effective way to equip the P-VLM with visual perception ability.
% We choose two of the most representative visual perception tasks, i.e., detection and segmentation.

\paragraph{Model Composition} We illustrate the framework of \ourmodel in Figure \ref{fig:framework}. \ourmodel mainly consists of an image-text aligned vision transformer~\cite{radford2021learning} (VIT) as image encoder, a large language model (LLM), and a set of lightweight visual perception encoders and decoders.
% , which encode visual perception signal as inputs and predicting them as outputs, respectively.
Given an image and a textual instruction, the image encoder first extracts the visual tokens from the image, which are then concatenated with text embeddings to serve as the input to the  pretrained LLM, such as Vicuna~\cite{vicuna2023}. Afterwards, the model is trained to autoregressively predict the next token. 

\paragraph{Empowering Visual Perception Capability} Rather than representing visual perception signals using discrete tokens as in previous approaches~\cite{peng2023kosmos2, chen2023shikra, wang2023visionllm, lu2022unifiedio, bai2023qwenvl, wang2022ofa}, which present difficulty during training, suffer from quantization error and increases the context length, we employ lightweight perception encoder and decoders to incorporate them into LLM's token embedding space. Specifically, we introduce a special token \loctoken, which indicates the presence of a visual perception signal. 

When perception signal serves as input, the lightweight encoders can map it to LLM's embedding space, then its embedding can be concatenated with other tokens before being processed by the LLM. During inference, if the token is classified as \loctoken, its corresponding embedding can be retrieved and passed to the light weight perception decoders, which then restores the perception signal to its original space. Specific to our implementation, the visual task decoder for segmentation mask estimation is implemented with two layers of two-way transformer block, in similar spirit as SAM~\cite{kirillov2023segment}. The mask encoder is a lightweight ResNet followed by a linear layer. The box encoder and decoder are both constructed with simple 3-layer MLPs. We leave the detailed description in the Appendix.

\paragraph{Training Objective} The utilization of LLM's token embedding allows us to harness the benefits of purposefully crafted training objectives for visual perception tasks, which take advantage of the prior knowledge and special characteristics inherent to the perception signals. Throughout the training process, we employ a combination of language modeling loss and task-specific losses. The overall training objective of \ourmodel during is: 
\begin{align}
    \mathcal{L_\text{all}}(S_{tar}, S_{in}, V_{tar}, I) &= \mathcal{L_\text{lang}}(S_{tar}, S_{in}, V_{tar}, I) \\
    + \mathcal{L_\text{vis}}(S_{in}, V_{tar}, I)
\end{align}
where $I$ is the input image, $S_{tar}, S_{in}, V_{tar}$ are  the target text, input instruction and target for visual perception signal, respectively.
In our case, since we select bounding box and segmentation mask prediction tasks to demonstrate our \ourmodel framework, our objective function can be formulated as:
\begin{align}
    \mathcal{L_\text{all}}(S_{tar}, S_{in}, \text{b}_\text{gt}, \text{m}_\text{gt}, I) = \mathcal{L_\text{lang}}(S_{tar}, S_{in}, I) \\+  \mathcal{L_\text{box}(\text{b}_\text{pred}, \text{b}_\text{gt})}
    +\mathcal{L_\text{mask}(\text{m}_\text{pred}, \text{m}_\text{gt})}
\end{align}
where $\text{b}_\text{pred}$ and $\text{b}_\text{gt}$ are the  prediction and ground truth bounding boxes, $\text{m}_\text{pred}$ and $\text{m}_\text{gt}$ are the  prediction and ground truth segmentation masks, respectively.

\textbf{Language Modelling Loss.} We adopt the autoregressive language modelling loss:
\begin{align}
    \mathcal{L_\text{lang}}(S_{tar}, S_{in}, I)=-\sum^L_{t=1}\log p\left[s^t_{tar} | \mathcal{F} (s^{(<t)}_{tar}, S_{in}, I)\right]
\end{align}

where $\mathcal{F}$ represents the VLM. $I$ represents the image, and $y_t$ denotes the $t^{th}$ token of the answer. $L$ is the length of the answer. $\mathcal{L_\text{lang}}$ supervises the model to generate corresponding output sentences based on the image and the input texts. In addition, it also teaches the model when to generate the \loctoken for predicting the perception signal.

\textbf{Bounding Box Loss.} We use a combination of L1-norm and GIoU loss for learning bounding box prediction:
\begin{align}
    \mathcal{L_\text{box}(\text{b}_\text{pred}, \text{b}_\text{gt})}= \lambda_{1}\mathcal{L_\text{L1}(\text{b}_\text{pred}, \text{b}_\text{gt})} \\ + \lambda_{2}\mathcal{L_\text{GIoU}(\text{b}_\text{pred}, \text{b}_\text{gt})}
\end{align}
% where $\text{b}_\text{pred}$ and $\text{b}_\text{gt}$ are the  prediction and ground truth bounding boxes, respectively.

\textbf{Mask Loss.} We use a combination of binary cross-entropy (BCE) loss DICE loss to enable \ourmodel to generate segmentation masks:
\begin{align}
    \mathcal{L_\text{mask}(\text{m}_\text{pred}, \text{m}_\text{gt})}=  \lambda_{3}\mathcal{L_\text{BCE}(\text{m}_\text{pred}, \text{m}_\text{gt})} \nonumber\\ + \lambda_{4}\mathcal{L_\text{DICE}(\text{m}_\text{pred}, \text{m}_\text{gt})}
\end{align}
% where $\text{m}_\text{pred}$ and $\text{m}_\text{gt}$ are the  prediction and ground truth bounding boxes, respectively. 
% The overall loss function of \ourmodel during training is: 
$\text{b}_\text{pred}$ and  $\text{m}_\text{pred}$ are derived by decoding from the corresponding \loctoken token embeddings via the visual perception decoders. In our experiments, we set all $\lambda_{(1-4)}$ to one, which demonstrates good results.
% The box loss $\mathcal{L_\text{box}}$ and mask loss $\mathcal{L_\text{mask}}$ are leveraged to enable \ourmodel to perform box and mask predictions, respectively. 
The use of visual task-specific training objectives not only boosts performance, but also makes training much easier and more efficient.
% than using only language modelling loss to learn such information in an auto-regressive manner. We further demonstrate this via experiment in Section~\ref{sec:dis_vs_cont}.

\paragraph{Multi-layer Visual Feature Fusion} Previous VLM approaches based on LLM predominantly depend on only the visual feature from last layer of pretrained VIT. However, this may be suboptimal for perception tasks, since the representations from top layers usually contain richer semantic features, while lacking fine-grained visual features. Inspired by layer-wise feature fusion technique in computer vision~\cite{lin2017feature}, we propose to make use of visual features from all layers of the VIT. Specifically, we learn an adaptive weighting term for each layer, and leverage the weighted-sum of representations across different layers:
\begin{align}
    Feat = \sum_{i=1}^{n} w_i \cdot Feat_i \quad \text{{subject to}} \quad \sum_{i=1}^{n} w_i = 1
\end{align}
where $Feat$ is the input image feature to the LLM, $w_i$ is the learnt weighting for image feature $Feat_i$ from $i^{th}$ layer.
% \paragraph{Remark} It is worth noting that enabling the model to process spatial input is one of the key advantages of end-to-end trained VLMs compared with two-stage ones, which makes it possible for the model to truly interpret the spatial concept, since those generated masks/boxes are still kept in the VLM's context.

\begin{table*}[htp!]
% \small
    \centering
 % \vspace{-0.25cm}
    % \resizebox{\textwidth}{!}{%
% \begin{small}
% \scalebox{0.9}
{
\begin{tabular}{c!{\vrule width 0.5pt}c!{\vrule width 0.5pt}ccc!{\vrule width 0.5pt}ccc!{\vrule width 0.5pt}cc}
% \begin{tabular}{p{1.3cm}!{\vrule width 0.5pt}p{1.3cm}p{1.3cm}p{1.3cm}!{\vrule width 0.5pt}p{1.3cm}p{1.3cm}p{1.3cm}!{\vrule width 0.5pt}p{1.3cm}p{1.3cm}p{1.3cm}}

\toprule
 & &\multicolumn{3}{c}{RefCOCO} & \multicolumn{3}{c}{RefCOCO+} & \multicolumn{2}{c}{RefCOCOg}\\
 % \cline{2-3} \cline{4-6} \cline{7-9}
Model Type&Method & val &  testA & testB &  val &  testA & testB &  val & test \\
\midrule
Two-stage&LISA~\citep{lai2023lisa}   &  74.9  & 79.1 & 72.3 & 65.1 & 70.8 & 58.1 & 67.9 & 70.6 \\
\midrule
\multirow{10}{*}{End-to-end}&MCN~\citep{luo2020multitask}     & 62.4 &  64.2 &  59.7 &  50.6 &  55.0 &  44.7 &  49.2 & 49.4\\
&VLT~\citep{ding2021visionlanguage}    &   67.5 & 70.5 & 65.2 & 56.3 & 61.0 & 50.1 & 55.0 & 57.7\\
&CRIS~\citep{wang2022cris}    &  70.5 & 73.2 & 66.1 & 62.3 & 68.1 & 53.7 & 59.9 & 60.4 \\
% \midrule
&LAVT~\citep{yang2022lavt}     &  72.7 & 75.8 & 68.8 & 62.1 & 68.4 & 55.1 & 61.2 & 62.1 \\
&ReLA~\citep{liu2023gres}    &  73.8 & 76.5 & 70.2 & 66.0 & 71.0 & 57.7 & 65.0 & 66.0\\
&X-Decoder~\citep{zou2022generalized}     &  -  & - & - & - & - & - & 64.6 & - \\
&SEEM~\citep{zou2023segment}   &  -  & - & - & - & - & - & 65.7 & - \\
\cmidrule(lr){2-10}
&\textbf{\ourmodel-7B}    &  \textbf{75.1} & \textbf{78.6} & \textbf{71.7} & \textbf{68.5} & \textbf{73.9} & \textbf{61.3} & \textbf{70.3} & \textbf{71.7}\\
&\textbf{\ourmodel-13B}   &  \textbf{75.3} & \textbf{79.1} & \textbf{72.1} & \textbf{68.9} & \textbf{74.0} & \textbf{61.9} & \textbf{70.7} & \textbf{71.9}\\
% \midrule
% G-DINO-L & 90.56 & 93.19 & 88.24 & 82.75 & 88.95 & 75.92 & 86.13 & 87.02\\
% UNINEXT-H & 92.64 & 94.33 & 91.46 & 85.24 & 89.63 & 79.79 & 88.73 & 89.37\\
\bottomrule
\end{tabular}
}
\caption{Results on refer segmentation (RES) task. Our \ourmodel significantly outperforms other end-to-end methods on all dataset splits, and also surpasses the two-stage approach LISA~\cite{lai2023lisa} by a large margin on majority of the splits.}
\label{tab:ref_seg}
% \end{small}
% }
% \vspace{-0.35cm}
\end{table*}

\begin{table}[htp!]
\small
\centering
{
\begin{tabular}{c|ccc}
\toprule
 % \multirow{2}{*}{\textbf{Size}}&\multicolumn{2}{c}{\textbf{IMDb}}  & \multicolumn{2}{c}{\textbf{Elec}} & \multicolumn{2}{c}{\textbf{ Yelp }}  \\
  \textit{\textsc{Method}}& \textit{\textsc{Quant Error Free}} & \textit{\textsc{Box}} & \textit{\textsc{Mask}}\\
\hline
 Shikra & \xmark & 20 & NA   \\
 Unified-IO & \xmark &  2 & 256 \\
 OFA & \xmark & 2 &  NA \\
 Kosmos-2 & \xmark & 2&  NA\\
 VisionLLM & \xmark & 2 & 16\\
 \ourmodel & \cmark & 1 & 1\\
% \hline
 % OFA & 60.01 & 61.69  \\
% \hline
% \multirow{2}{*}{Robin} &  7B & 56.32 & 60.21 & 62.73  & \cc{59.75} \\
% & 13B & 61.03 & 61.29 & 64.36 & \cc{62.23}  \\
% & &   22.78 & 24.97 & 27.64 & \cc{25.13} \\
%  & & \textbf{23.62} & \textbf{28.27}&\textbf{ 31.50}& \cc{\textbf{27.80}} \\
%  &  & 15.16 & 18.05 & 22.16 & \cc{18.46} \\ 
%  & & 23.04 & 27.98 & 30.11 & \cc{27.04} \\
\bottomrule
\end{tabular}}
\caption{Number of tokens needed to represent boxes or segmentation mask. Our \ourmodel is able to represent both box and mask with only one token, while prevents quantization error.}
\label{tab:token_number}
\end{table}
\section{Experiments}
% \begin{table}[t]
% \centering
% {
% \begin{tabular}{c|cccccc}
% \toprule
%  % \multirow{2}{*}{\textbf{Size}}&\multicolumn{2}{c}{\textbf{IMDb}}  & \multicolumn{2}{c}{\textbf{Elec}} & \multicolumn{2}{c}{\textbf{ Yelp }}  \\
%   & \textit{\textsc{OFA}}& \textit{\textsc{Unified-IO}} & \textit{\textsc{Shikra}} &\textit{\textsc{Kosmos-2}} & \textit{\textsc{VisionLLM}} &\textit{\textsc{\ourmodel}}\\
% \hline
%  % Shikra & 10.27 & 12.30   \\
%  % Unified-IO & 7B &  59.45 \\
%  % OFA & 60.01 & 61.69  \\
%  % Kosmos-2 & 12.53 & 13.94   \\
%  VisionLLM & 7B & {61.11}\\
% % \hline
%  % OFA & 60.01 & 61.69  \\
% % \hline
% % \multirow{2}{*}{Robin} &  7B & 56.32 & 60.21 & 62.73  & \cc{59.75} \\
% % & 13B & 61.03 & 61.29 & 64.36 & \cc{62.23}  \\
% % & &   22.78 & 24.97 & 27.64 & \cc{25.13} \\
% %  & & \textbf{23.62} & \textbf{28.27}&\textbf{ 31.50}& \cc{\textbf{27.80}} \\
% %  &  & 15.16 & 18.05 & 22.16 & \cc{18.46} \\ 
% %  & & 23.04 & 27.98 & 30.11 & \cc{27.04} \\
% \bottomrule
% \end{tabular}}
% \caption{Performance with different language models.}
% \label{tab:token_number}
% \end{table}

\subsection{Training and Evaluation}
\paragraph{Datasets} Similar as in Shikra~\cite{chen2023shikra}, to equip \ourmodel with visual perception ability, we adopt RefCOCO~\cite{yu2016modeling}, RefCOCO+~\cite{yu2016modeling}, RefCOCOg~\cite{mao2016generation}, Visual Gemone~\cite{krishna2016visual} and Flicker30k~\cite{plummer2016flickr30k}. Since Visual Genome and Flicker30k do not have segmentation mask annotations, we leverage the powerful SAM~\cite{kirillov2023segment} as an auto-labelling system to generate masks from bounding box annotations. For captioning, we leverage the COCO~\cite{lin2015microsoft} dataset and the image caption data in curated by LLAVA~\cite{liu2023llava}.
\paragraph{Hyperparameters} 
% Throughout our experiments, 
If not otherwise specified,  we use the following hyper-parameters throughtout all experiments: We initialize the LLM component with Vicuna~\cite{vicuna2023} weights, we adopt LoRA with rank set to 32, the learning rate is set to 3e-4, the batch size is 32 on each GPU during training. We run experiments on 8 A40 GPUs with 80G memory for 70 hours in total. For ablation study, we use Vicuna-7B as the LLM backbone to conduct experiments.
\subsection{Qualitative Results}
We demonstrate the capability of our \ourmodel with some generated results in Figure~\ref{fig:perceptiongpt_demo}, which showcases the following capabilities: (1) Spotting Captioning (first 4 rows), which is capable of generating the captions while spoting the objects with boxes and segmentation masks. (2) Reasoning-based detection and Segmentation (row 4-8). (3) Image-level and region-level captioning (row 9). We surprisingly find that our model is able to perform reasoning based on the image and the user query, then detect and segment the object of interest, even though such ability is not specifically considered in the training data. 

\subsection{Quantitive Results}

We show the performance of \ourmodel by evaluating it on a series of benchmarks, which demonstrates competitive results across a variety of tasks.

\paragraph{Refering Expression Comprehension (REC)} The REC task mainly aims to understand the image and a textual phrase, and  then localize the referred object by drawing a bounding box around it. We compare our \ourmodel with both generalist and specialist approaches in Table \ref{tab:ref_det}. Notably, \ourmodel achieves SOTA performances with only parameter-efficient tuning. We further verify that encoding visual perception signals into token embeddings of LLM is the key to reducing training difficulty in Section \ref{sec:dis_vs_cont}.

\paragraph{Refering Expression Segmentation (RES)} Compared with the REC task, the RES task requires image-text understanding at the finer pixel level, which requires the prediction of segmentation mask that seperates the referred object from the image. Owing to the use of token embedding to encode the visual perception signals, our method is able to effectively handle segmentation masks with a single token embedding, which would have required dozens of discrete tokens to represent in previous P-VLM methods. We compare our \ourmodel with other approaches in Table \ref{tab:ref_seg}. Our method achieves SOTA performance on this task, with only parameter-efficient training and without relying on external vision experts.
% We wish to note that there is room for improvement, since one bottleneck of the performance of \ourmodel is the input resolution from the visual backbones, which is not the focus of our paper and can be expected to be resolved in future work.

\paragraph{Conventional Vision-language Tasks} We evaluate \ourmodel's ability on conventional vision-language tasks, namely image captioning (IC) and visual question answering (VQA). We finetune our \ourmodel on the training split of the datasets before evaluation and show the results for comparision in Table \ref{tab:cap_vqa}, which demonstrate that our method is comparable with specialized and generalist models at conventional image-language tasks. The superiority of our method compared with Shikra may be attributed to the use of token embedding to represent perception signal, which eases learning and alleviates forgetting.

\begin{table*}
    \centering
    \begin{tabular}{l|cccccccc}
        \toprule
        Datasets & PerceptGPT & Shikra & FM-80B & FM-9B & Kosmos-1 & BLIP-2 & Unified-IO & VPGTrans \\
        \midrule
        VQAv2 & \textbf{85.1} & 83.3 & 56.3 & 51.8 & 51.0 & 65.2 & 77.9 & 65.2 \\
        OK-VQA & \textbf{56.2} & 53.8 & 50.6 & 44.7 & - & 45.9 & 54.0 & 45.0 \\
        \midrule
        Flickr30k & \textbf{77.1} & 73.9 & - & - & 67.1 & - & - & - \\
        COCO & \textbf{123.2} & 117.5 & 84.3 & 79.4 & 84.7 & - & 122.1 & 114.2 \\
        \bottomrule
    \end{tabular}
    \caption{Comparison on VQA and Image Captioning tasks. For VQA, conduct evaluation on VQAv2 \cite{goyal2017making} and OK-VQA \cite{marino2019okvqa} using Accuracy (\%) as the metric. For Image Captioning, we evaluate them on COCO \cite{lin2015microsoft} and Flickr30k \cite{plummer2016flickr30k} using CIDEr.}
    \label{tab:cap_vqa}
\end{table*}

\subsection{Ablation Study}
\paragraph{\ourmodel Alleviates Learning Difficulty}\label{sec:dis_vs_cont}
Leveraging LLM's token embedding to represent visual perception signals enables training to be efficient in terms of both parameters and data. As shown in Table~\ref{tab:tokenization}, we train the models with only the concatenated RefCOCO, RefCOCO+ and RefCOCOg training splits, and evaluate the Acc@0.5 for bounding box predictions. We observe that numerical representation does not enable parameter-efficient training (LoRA), while our \ourmodel performs well even with LoRA rank set to 8. In addition, the performance of discrete representation declines greatly without the large-scale visual genome dataset, while ours still remains promising. 

The above phenomenon may be attributed to the following reasons: 1) the sequence representation for bounding boxes are not optimal, since the coordinate values do not have causal relationships as in NLP. On the other hand, directly decoding from continuous embedding does not need to model the dependencies between discrete tokens; 2) continuous embeddings encapsulate rich, multidimensional data in a dense form, allowing for a more compact and efficient representation of spatial information than a sequence of discrete tokens; 3) those discrete representations can not leverage the specially designed loss functions for vision tasks as in our continuous representation;
\begin{table*}
    \centering

    \begin{tabular}{lccccccccc}
        \toprule
        & &\multicolumn{3}{c}{RefCOCO} & \multicolumn{3}{c}{RefCOCO+} & \multicolumn{2}{c}{RefCOCOg} \\
        \cmidrule(lr){3-5} \cmidrule(lr){6-8} \cmidrule(lr){9-10}
        Method & & val & testA & testB & val & testA & testB & val & test \\
        \midrule
        \multirow{4}{*}{Numerical}&\quad 8 & 0.79& 0.63 & 0.51 & 0.28 & 0.36 & 0.25 & 0.43 & 0.41 \\
        % &\quad 16 & - & - & - & - & - & - & - \\
        &\quad 32 & 10.5& 10.1 & 9.64 & 8.72 & 10.4 & 9.27 & 8.63& 10.2 \\
        &\quad full & 45.7& 53.6 & 49.2 & 43.5 & 44.1 & 39.7 & 42.1 & 41.3 \\
        \midrule
        % \multirow{4}{*}{Vocab}&\quad 8 & -& - & - & - & - & - & - & - \\
        % % &\quad 16 & - & - & - & - & - & - & - \\
        % &\quad 32 & -& - & - & - & - & - & - & - \\
        % &\quad full& - & - & - & - & - & - & - & - \\
        % \midrule
        \multirow{4}{*}{Continuous}&\quad 8 & \textbf{69.7}& \textbf{73.2} & \textbf{72.5} & \textbf{68.4 }& \textbf{70.3} &\textbf{ 68.1} & \textbf{71.6} & \textbf{70.5}\\
        % &\quad 16 & - & - & - & - & - & - & - \\
        &\quad 32 &  \textbf{71.2} & \textbf{75.4} & \textbf{74.1} & \textbf{70.5} & \textbf{71.9} & \textbf{69.6} & \textbf{72.5} & \textbf{72.9}\\
        &\quad full& \textbf{75.5} & \textbf{79.6} & \textbf{78.1} & \textbf{74.9} & \textbf{74.4} & \textbf{72.1} & \textbf{76.4} & \textbf{75.2}\\
        \bottomrule
    \end{tabular}
    \caption{Experiment on the influence of different tokenization techniques.  We observe that neither numerical or vocabulary representations perform well with LoRA training.  On the other hand, using LLM's token embedding to decode bounding boxes  achieves good performance even with low LoRA ranks.}
    \label{tab:tokenization}
\end{table*}
% \paragraph{The Impact of Lora Rank}
\paragraph{Impact of Layer Fusion}
% \begin{figure}
%   \centering
%   \includegraphics[width=0.4\textwidth]{figures/bar_plot.png}
%   \caption{This is an example figure.}
%   \label{fig:weight_distribution}
% \end{figure}
\begin{figure}[htbp]
  \begin{subfigure}[b]{0.48\linewidth}
    \includegraphics[width=\linewidth]{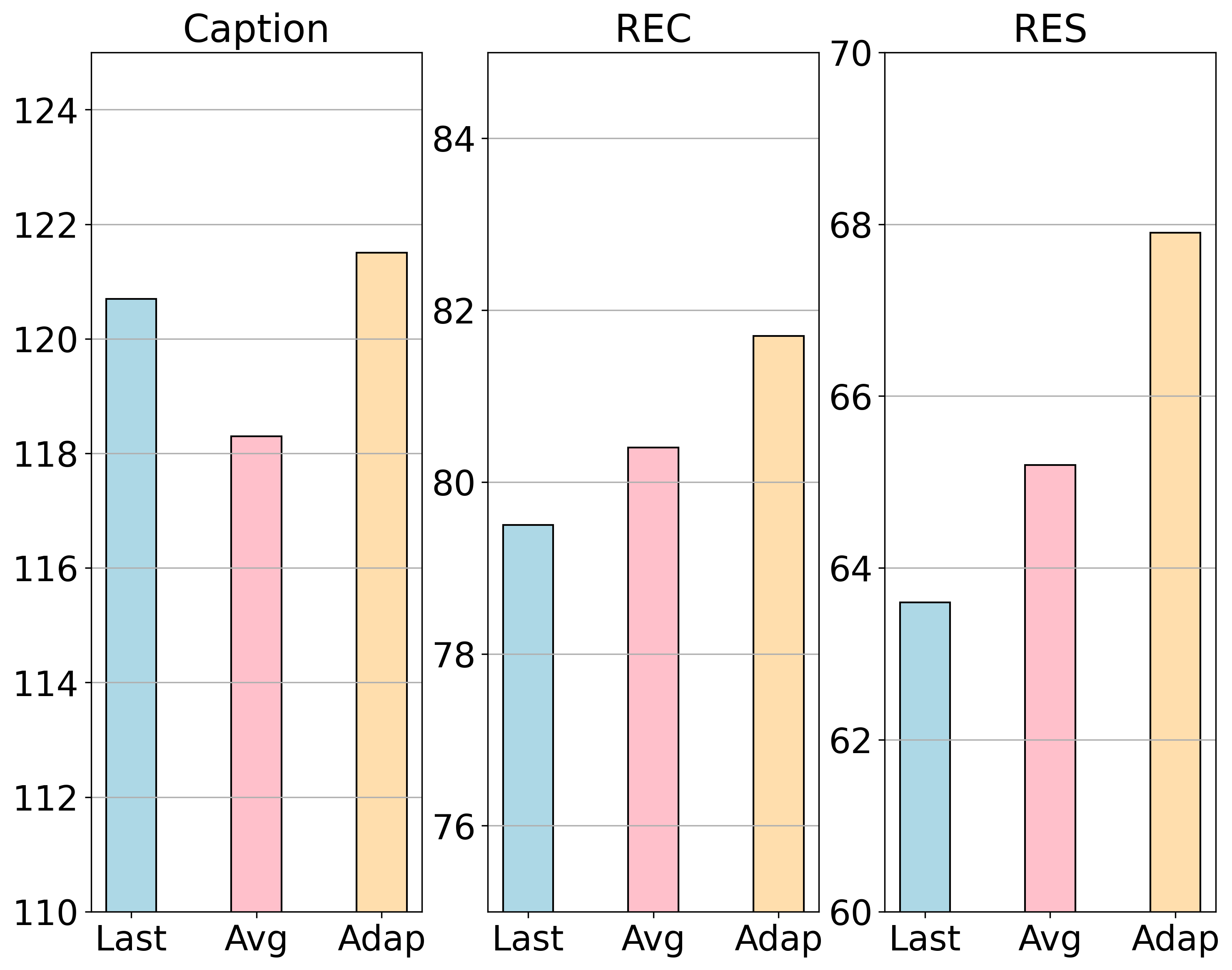}
    % \caption{Weight distribution across different layers of visual features.}
  \end{subfigure}
  \hfill
  \begin{subfigure}[b]{0.48\linewidth}
    \includegraphics[width=\linewidth]{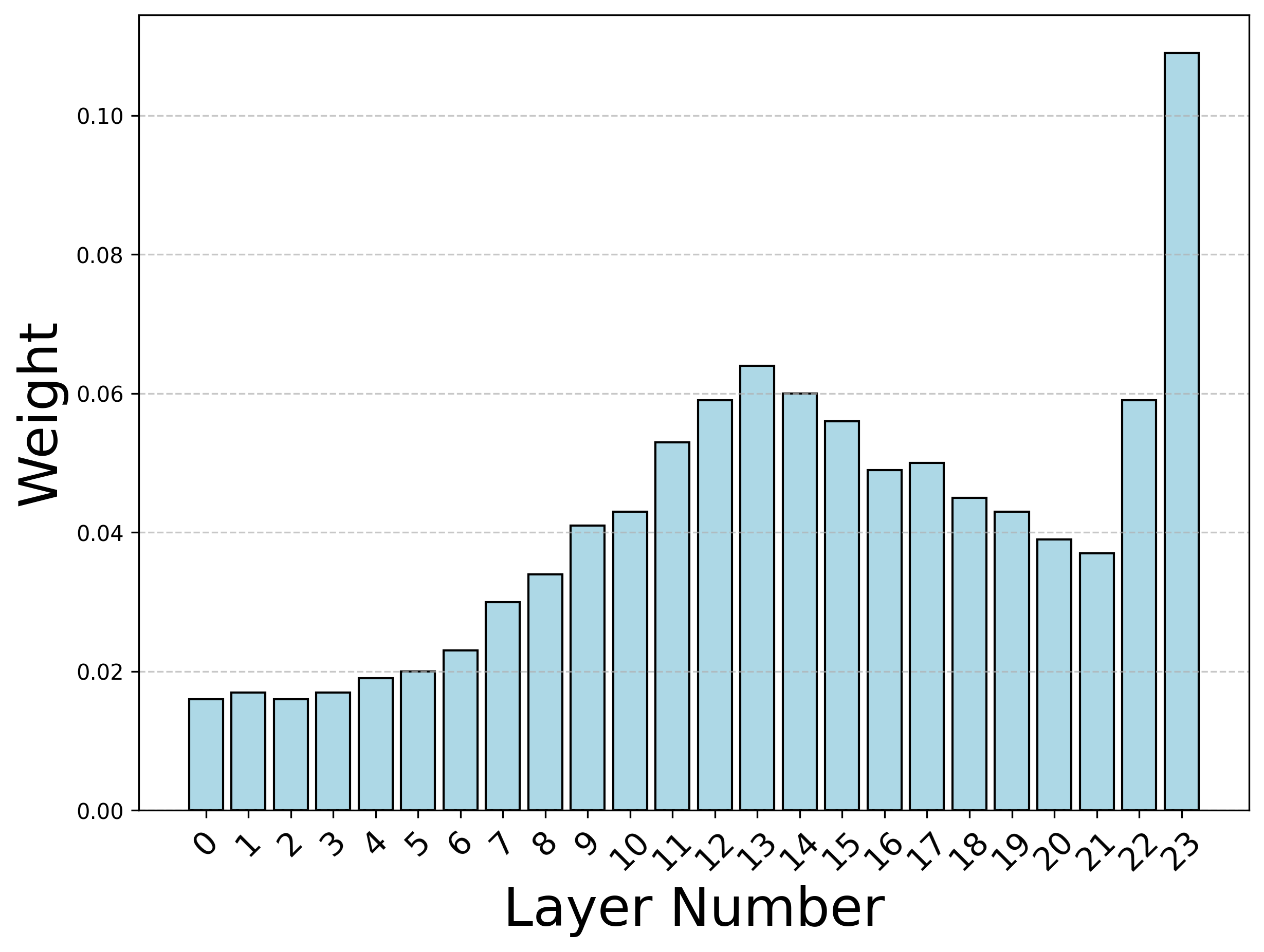}
    % \caption{Weight distribution across different layers of visual features.}
  \end{subfigure}
  % \begin{subfigure}[b]{0.48\linewidth}
  %   \centering
  %   \small
  %   \captionsetup{skip=30pt}
  %   \begin{tabular}{cccc}
  %     \hline
  %      & Cap & REC & RES \\
  %     \hline
  %     Last & 120.7 & 79.5 & 63.6\\
  %     Avg & 118.3 & 80.4 & 65.2\\
  %     Ada & 121.5 & 81.7 & 67.9\\
  %     \hline
  %   \end{tabular}
  %   % \caption{This is an example table.}
  %   \label{tab:fusion_strategy}
  %   \vspace{25pt}
  % \end{subfigure}
  \vspace{-10pt}
  \caption{Left: The performance of different strategy for fusing visual features on various tasks. Right: The magnitude of learnt adaptive weights for visual features across different VIT layers.}
  \label{fig:vit_fusion}
\end{figure}
We examine the impact of different strategies for deriving the visual features from VIT. As shown in left sided of Figure~\ref{fig:vit_fusion}, the features from top layers are more important for tasks such as image captioning, since they contain richer high-level semantic information of the image. On the other hand, features from bottom layers are more helpful for fine-grained visual perception tasks. With our adaptive fusion strategy, the weight for each layer's visual feature can be automatically adjusted. At the right side of the figure, we demonstrate the learnt weight  distribution across all layers. This simple yet effective strategy boosts the performance of \ourmodel on all tasks.

\begin{table}[htp!]
\centering
{
\begin{tabular}{c|cccc}
\toprule
 % \multirow{2}{*}{\textbf{Size}}&\multicolumn{2}{c}{\textbf{IMDb}}  & \multicolumn{2}{c}{\textbf{Elec}} & \multicolumn{2}{c}{\textbf{ Yelp }}  \\
 &\multicolumn{2}{c}{7b}&\multicolumn{2}{c}{13b}\\
  \textit{\textsc{Method}}& box & mask& box & mask\\
\hline
 Numerical & 3.62 & 42.1& 4.30 & 57.1    \\
 Vocab  &  0.26 & 4.01 & 0.45 & 6.16  \\
 Token Embed & 0.15 & 0.15 & 0.21 & 0.21  \\
% \hline
 % OFA & 60.01 & 61.69  \\
% \hline
% \multirow{2}{*}{Robin} &  7B & 56.32 & 60.21 & 62.73  & \cc{59.75} \\
% & 13B & 61.03 & 61.29 & 64.36 & \cc{62.23}  \\
% & &   22.78 & 24.97 & 27.64 & \cc{25.13} \\
%  & & \textbf{23.62} & \textbf{28.27}&\textbf{ 31.50}& \cc{\textbf{27.80}} \\
%  &  & 15.16 & 18.05 & 22.16 & \cc{18.46} \\ 
%  & & 23.04 & 27.98 & 30.11 & \cc{27.04} \\
\bottomrule
\end{tabular}}
\caption{The inference time taken to decode a box or a mask under different formulations. In our approach, the visual perception information can be contained in a single token embedding, which greatly boosts the inference efficiency.}
\label{tab:inference_speed}
\end{table}

\paragraph{Inference Speed Comparison}
We compare the inference speed between different formulations of visual perception information in Table~\ref{tab:inference_speed}. The number of points to represent a mask is 32, which ensures acceptable mask quality. Since \ourmodel requires only one token embedding to carry the visual perception information, the inference speed greatly boosted, especially for lengthy information such as segmentation masks. Specifically, for a 7B P-VLM to decode a mask, \ourmodel takes only 0.3\% and 3.7\% inference time of Numerical and Vocabulary formulations.

% \paragraph{The Impact of Lora Rank}
\section{Conclusion} 
In this paper, we propose \textbf{\ourmodel}, a novel framework for efficiently training perception-enhanced vision language models (P-VLMs). Our approach addresses the limitations of existing methods by taking advantage of the representation power of the LLM's token embeddings. Our framework achieves promising results by tuning only a small fraction of parameters, resulting in compact representations and significantly accelerated decoding. We hope this work provides new insights into future research of P-VLMs. 
{
    \small
    \bibliographystyle{ieeenat_fullname}
    \bibliography{main}
}

% WARNING: do not forget to delete the supplementary pages from your submission 
% \input{sec/X_suppl}

\end{document}